# Joint Representation Learning of Cross-lingual Words and Entities via Attentive Distant Supervision

**Yixin Cao**[1,2] **Lei Hou**[2*] **Juanzi Li**[2] **Zhiyuan Liu**[2]
**Chengjiang Li**[2] **Xu Chen**[2] **Tiansi Dong**[3]
[1]School of Computing, National University of Singapore, Singapore
[2]Department of CST, Tsinghua University, Beijing, China
[3]B-IT, University of Bonn, Bonn, Germany
{caoyixin2011,iamlockelightning,successcx}@gmail.com
{houlei,liuzy,lijuanzi}@tsinghua.edu.cn
dongt@bit.uni-bonn.de

## Abstract

Joint representation learning of words and entities benefits many NLP tasks, but has not been well explored in cross-lingual settings. In this paper, we propose a novel method for joint representation learning of cross-lingual words and entities. It captures mutually complementary knowledge, and enables cross-lingual inferences among knowledge bases and texts. Our method does not require parallel corpora, and automatically generates comparable data via distant supervision using multi-lingual knowledge bases. We utilize two types of regularizers to align cross-lingual words and entities, and design knowledge attention and cross-lingual attention to further reduce noises. We conducted a series of experiments on three tasks: word translation, entity relatedness, and cross-lingual entity linking. The results, both qualitatively and quantitatively, demonstrate the significance of our method.

## 1 Introduction

Multi-lingual knowledge bases (KB) store millions of entities and facts in various languages, and provide rich background structural knowledge for understanding texts. On the other hand, text corpus contains huge amount of statistical information complementary to KBs. Many researchers leverage both types of resources to improve various natural language processing (NLP) tasks, such as machine reading (Yang and Mitchell, 2017), question answering (He et al., 2017; Hao et al., 2017).

Most existing work jointly models KB and text corpus to enhance each other by learning word and entity representations in a unified vector space. For example, Wang et al. (2014); Yamada et al. (2016); Cao et al. (2017) utilize the co-occurrence information to align similar words and entities with similar embedding vectors. Toutanova et al. (2015); Wu et al. (2016); Han et al. (2016); Weston et al. (2013a); Wang and Li (2016) represent entities based on their textual descriptions together with the structured relations. These methods focused on mono-lingual settings. However, for cross-lingual tasks (e.g., cross-lingual entity linking), these approaches need to introduce additional tools to do translation, which suffers from extra costs and inevitable errors (Ji et al., 2015, 2016).

In this paper, we carry out cross-lingual joint representation learning, which has not been fully researched in the literature. We aim at creating a unified space for words and entities in various languages, and easing cross-lingual semantic comparison, which will benefit from the complementary information in different languages. For instance, two different meanings of word *center* in English are expressed by two different words in Chinese: *center* as *the activity-specific building* is expressed by 中心, *center* as *the basketball player role* is 中锋.

Our main challenge is the limited availability of parallel corpus, which is usually either expensive to obtain, or only available for certain narrow domains (Gouws et al., 2015). Many work has been done to alleviate the problem. One school of methods uses adversarial technique or domain adaption to match linguistic distribution (Zhang et al., 2017b; Barone, 2016; Cao et al., 2016). These methods do not require parallel corpora. The weakness is that the training process is unstable and that the high complexity restricts the methods only to small-scale data. Another line of work uses pre-existing multi-lingual resources to automatically generate "pseudo bilingual documents" (Vulic and Moens, 2015, 2016). However, negative results have been observed due to the occasional poor quality of training data (Vulic and Moens, 2016). All above methods only focus on words. We consider both words and entities, which makes the parallel data issue more challenging.

*Corresponding author.

In this paper, we propose a novel method for joint representation learning of cross-lingual words and entities. The basic idea is to capture mutually complementary knowledge in a shared semantic space, which enables joint inference among cross-lingual knowledge base and texts without additional translations. We achieve it by (1) utilizing an existing multi-lingual knowledge base to automatically generate cross-lingual supervision data, (2) learning mono-lingual word and entity representations, (3) applying cross-lingual sentence regularizer and cross-lingual entity regularizer to align similar words and entities with similar embeddings. The entire framework is trained using a unified objective function, which is efficient and applicable to arbitrary language pairs that exist in multi-lingual KBs.

Particularly, we build a bilingual entity network from inter-language links [1] in KBs for regularizing cross-lingual entities through a variant of skip-gram model (Mikolov et al., 2013c). Thus, mono-lingual structured knowledge of entities are not only extended to cross-lingual settings, but also augmented from other languages. On the other hand, we utilize **distant supervision** to generate comparable sentences for cross-lingual sentence regularizer to model co-occurrence information across languages. Compared with "pseudo bilingual documents", comparable sentences achieve higher quality, because they rely not only on the shared semantics at document level, but also on cross-lingual information at sentence level. We further introduce two attention mechanisms, knowledge attention and cross-lingual attention, to select informative data in comparable sentences.

Our contributions can be concluded as follows:

- We proposed a novel method that jointly learns representations of not only cross-lingual words but also cross-lingual entities in a unified vector space, aiming to enhance the embedding quality from each other via complementary semantics.

- Our proposed model introduces distant supervision coupled with attention mechanisms to generate comparable data as cross-lingual supervision, which can benefit many cross-lingual analysis.

- We did qualitative analysis to have an intuitive impression of our embeddings, and quantitative analysis in three tasks: word translation, entity relatedness, and cross-lingual entity linking. Experiment results show that our method demonstrates significant improvements in all three tasks.

## 2 Related Work

Jointly representation learning of words and entities attracts much attention in the fields of Entity Linking (Zhang et al., 2017a; Cao et al., 2018), Relation Extraction (Weston et al., 2013b) and so on, yet little work focuses on cross-lingual settings. Inspiringly, we investigate the task of cross-lingual word embedding models (Ruder et al., 2017), and classify them into three groups according to parallel corpora used as supervisions: (i) methods requiring **parallel corpus with aligned words** as constraint for bilingual word embedding learning (Klementiev et al., 2012; Zou et al., 2013; Wu et al., 2014; Luong et al., 2015; Ammar et al., 2016; Soricut and Ding, 2016). (ii) methods using **parallel sentences** (i.e. translated sentence pairs) as the semantic composition of multi-lingual words (Gouws et al., 2015; Kociský et al., 2014; Hermann and Blunsom, 2014; Chandar et al., 2014; Shi et al., 2015; Mogadala and Rettinger, 2016). (iii) methods requiring **bilingual lexicon** to map words from one language into the other (Mikolov et al., 2013b; Faruqui and Dyer, 2014; Xiao and Guo, 2014).

The major weakness of these methods is the limited availability of parallel corpora. One remedy is to use existing multi-lingual resources (i.e. multi-lingual KB). Camacho-Collados et al. (2015) combines several KBs (Wikipedia, WordNet and BabelNet) and leverages multi-lingual synsets to learn word embeddings at sense level through an extra post-processing step. Artetxe et al. (2017) starts from a small bilingual lexicon and using a self-learning approach to induce the structural similarity of embedding spaces. Vulic and Moens (2015, 2016) collect comparable documents on same themes from multi-lingual Wikipedia, shuffle and merge them to build "pseudo bilingual documents" as training corpora. However, the quality of "pseudo bilingual documents" are difficult to control, resulting in poor performance in several cross-lingual tasks (Vulic and Moens, 2016).

Another remedy matches linguistic distribu-tion via adversarial training (Barone, 2016; Zhang et al., 2017b; Lample et al., 2018), domain adaption (Cao et al., 2016). However, these methods suffer from the instability of training process and the high complexity. This either limits the scalability of vocabulary size or relies on a strong distribution assumption.

[1] https://en.wikipedia.org/wiki/Help:Interlanguage_links

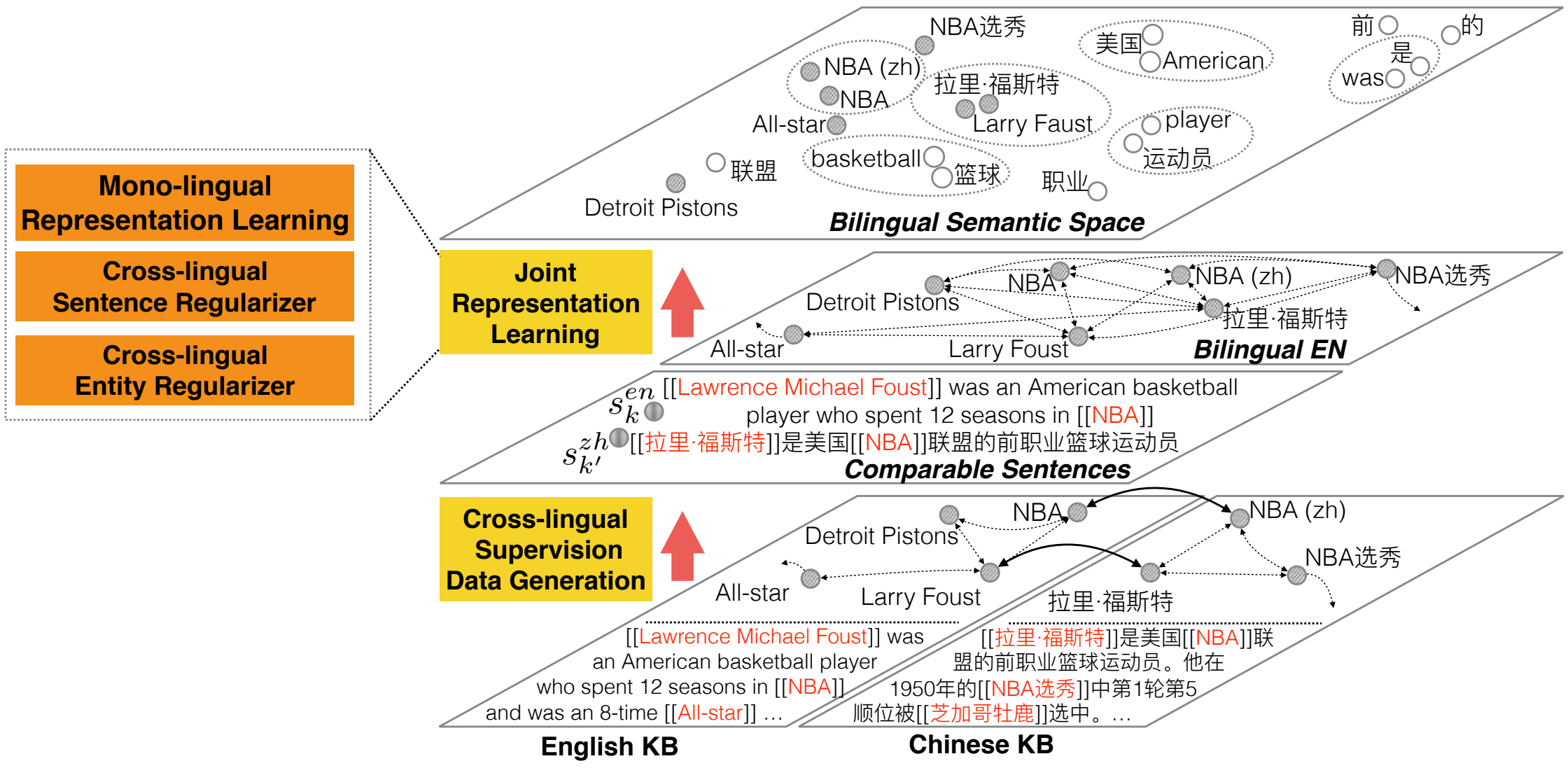

Figure 1: The overview framework of our method. The inputs and outputs of each step are listed in the three levels. Particularly, there are three main components of joint representation learning. Red texts with brackets are anchors, dashed lines denote entity relations, and solid lines are cross-lingual links.

Inspired by Vulic and Moens (2016), we generate highly qualified **comparable sentences** via distant supervision, which is one of the most promising approaches to addressing the issue of sparse training data, and performs well in relation extraction (Lin et al., 2017a; Mintz et al., 2009; Zeng et al., 2015; Hoffmann et al., 2011; Surdeanu et al., 2012). Our comparable sentences may further benefit many other cross-lingual analysis, such as information retrieval (Dong et al., 2014).

# 3 Preliminaries and Framework

## 3.1 Preliminaries

Given a multi-lingual KB, we take (i) text corpus, (ii) entity and their relations, (iii) a set of anchors as inputs, and learn embeddings for each word and each entity in various languages. For clarity, we use English and Chinese as sample languages in the rest of the paper, and use superscript $y \in \{en, zh\}$ to denote language-specific parameters[2].

We use multi-lingual Wikipedia as KB including a set of entities $\mathcal{E}^y = \{e_i^y\}$ and their articles. We concatenate these articles together, and form text corpus $\mathcal{D}^y = \langle w_1^y, \dots, w_i^y, \dots, w_{|\mathcal{D}|}^y \rangle$. Hyper links in articles are denoted by Anchors $\mathcal{A}^y = \{\langle w_i^y, e_j^y \rangle\}$, which indicates that word $w_i^y$ refers to entity $e_j^y$. $\mathcal{G}^y = (\mathcal{E}^y, \mathcal{R}^y)$ is the mono-lingual Entity Network (EN), where $\mathcal{R}^y = \{\langle e_i^y, e_j^y \rangle\}$ if there is a link between $e_i^y, e_j^y$. We use inter-language links in Wikipedia as cross-lingual links $\mathcal{R}^{en-zh} = \{\langle e_i^{en}, e_{i'}^{zh} \rangle\}$, indicating $e_i^{en}$, $e_{i'}^{zh}$ refer to the same thing in English and Chinese. **Cross-lingual word and entity representation learning** is to map words and entities in different languages into a unified semantic space. Each word and entity obtain their embedding vectors[3] $\mathbf{w}_i^y$ and $\mathbf{e}_j^y$.

## 3.2 Framework

To alleviate the heavy burden of limited parallel corpora and additional translation efforts, we utilize existing multi-lingual resources to distantly supervise cross-lingual word and entity representation learning, so that the shared embedding space supports joint inference among KB and texts across languages. As shown in Figure 1, our framework has two steps: (1) **Cross-lingual Su-**

[2]We choose English and Chinese as example languages because they are top-ranked according to total number of speakers, the full list can be found in https://en.wikipedia.org/wiki/Lists_of_languages_by_number_of_speakers.

[3]For the cross-lingual linked entities sharing the same strings (e.g., *NBA* and *NBA (zh)*), which is an infrequent situation between languages, we use separated representations to keep training objective consistent and avoid confusion.

**pervision Data Generation** builds a bilingual entity network and generates comparable sentences based on cross-lingual links; (2) **Joint Representation Learning** learns cross-lingual word and entity embeddings using a unified objective function. Our assumption throughout the entire framework is as follows: *The more words/entities two contexts share, the more similar they are.*

As shown in Figure 1, we build a bilingual EN $\mathcal{G}^{en-zh}$ by using $\mathcal{G}^{en}, \mathcal{G}^{zh}$ and cross-lingual links $\mathcal{R}^{en-zh}$. Thus, entities in different languages shall be connected in a unified network to facilitate cross-lingual entity alignments. Meanwhile, from KB articles, we extract comparable sentences $\mathcal{S}^{en-zh} = \{\langle s_k^{en}, s_k^{zh}\rangle\}$ as high qualified parallel data to align similar words in different languages.

Based on generated cross-lingual data $\mathcal{G}^{en-zh}, \mathcal{S}^{en-zh}$ and mono-lingual data $\mathcal{D}^y$, $\mathcal{A}^y$, where $y \in \{en, zh\}$, we jointly learn cross-lingual word and entity embeddings through three components: (1) **Mono-lingual Representation Learning**, which learns mono-lingual word and entity embeddings for each language by modeling co-occurrence information through a variant of skip-gram model (Mikolov et al., 2013c). (2) **Cross-lingual Entity Regularizer**, which aligns entities that refer to the same thing in different languages by extending the mono-lingual model to bilingual EN. For example, entity *Foust* in English and entity 福斯特 *(Foust)* in Chinese are closely embedded in the semantic space because they share common neighbors in two languages, *All-star* and *NBA* 选秀 *(draft)*, etc.. (3) **Cross-lingual Sentence Regularizer**, which models cross-lingual co-occurrence at sentence level in order to learn translated words to have most similar embeddings. For example, English word *basketball* and the translated Chinese word 篮球 frequently co-occur in a pair of comparable sentences, therefore, their vector representations shall be close in the semantic space. The above components are trained jointly under a unified objective function.

# 4 Cross-lingual Supervision Data Generation

This section introduces how to build a bilingual entity network $\mathcal{G}^{en-zh}$ and comparable sentences $\mathcal{S}^{en-zh}$ from a multi-lingual KB.

## 4.1 Bilingual Entity Network Construction

Entities with cross-lingual links refer to the same thing, which implies they are equivalent across languages. Conventional knowledge representation methods only add edges between $e_i^{en}$ and $e_{i'}^{zh}$ indicating a special "equivalent" relation (Zhu et al., 2017). Instead, we build $\mathcal{G}^{en-zh} = (\mathcal{E}^{en} \cup \mathcal{E}^{zh}, \mathcal{R}^{en} \cup \mathcal{R}^{zh} \cup \tilde{\mathcal{R}}^{en-zh})$ by enriching the neighbors of cross-lingual linked entities. That is, we add edges $\tilde{\mathcal{R}}^{en-zh}$ between two mono-lingual ENs by letting all neighbors of $e_i^{en}$ be neighbors of $e_{i'}^{zh}$, and vice versa, if $\langle e_i^{en}, e_{i'}^{zh}\rangle \in \mathcal{R}^{en-zh}$.

$\mathcal{G}^{en-zh}$ extends $\mathcal{G}^{en}$ and $\mathcal{G}^{zh}$ to bilingual settings in a natural way. It not only keeps a consistent objective in mono-lingual ENs—entities, no matter in which language, will be embedded closely if share common neighbors—but also enhances each other with more neighbors in the foreign language.

Following the method in Zhu et al. (2017), there will be no edge between Chinese entity 福斯特 *(Foust)* and English entity *Pistons*, which implies a wrong fact that 福斯特 *(Foust)* does not belong to *Pistons*. Our method enriches the missing relation between entities 福斯特 *(Foust)* and 活塞队 *(Pistons)* in incomplete Chinese KB through corresponding English common neighbors, *Allstar*, *NBA*, etc., as illustrated in Figure 1.

## 4.2 Comparable Sentences Generation

To supervise the cross-lingual representation learning of words, we automatically generate comparable sentences as cross-lingual training data. Comparable sentences are not translated paired sentences, but sentences with the same topic in different languages. As shown in the middle layer (Figure 1), the pair of sentences are comparable sentences: (1) "*Lawrence Michael Foust was an American basketball player who spent 12 seasons in NBA*", (2) "拉里·福斯特 *(Lawrence Foust)* 是 *(was)* 美国 *(American)* *NBA* 联盟 *(association)* 的 *(of)* 前 *(former)* 职业 *(professional)* 篮球 *(basketball)* 运动员 *(player)*".

Inspired by the distant supervision technique in relation extraction, we assume that sentence $s_k^{en}$ in Wikipedia articles of entity $e_i^{en}$ explicitly or implicitly describes $e_i^{en}$ (Yamada et al., 2017), and that $s_k^{en}$ shall express a relation between $e_i^{en}$ and $e_j^{en}$ if another entity $e_j^{en}$ is in $s_k^{en}$. Meanwhile, we find a comparable sentence $s_{k'}^{zh}$ in another language which satisfies $s_{k'}^{zh}$ containing $e_{j'}^{zh}$ in Wikipedia articles of Chinese entity $e_{i'}^{zh}$, where $\langle e_i^{en}, e_{i'}^{zh} \rangle, \langle e_j^{en}, e_{j'}^{zh} \rangle \in \mathcal{R}^{en-zh}$. As shown in Figure 1, the sentences in the second level are comparable due to the similar theme of the relation between entity *Foust* and *NBA*. To find this type of sentences, we search the anchors in the English aritcle and Chinese article of cross-lingual entity *Foust*, respectively, and extract the sentences including another crosslingual entity *NBA*. Comparable sentences can be regarded as cross-lingual contexts.

Unfortunately, comparable sentences suffer from two issues caused by distant supervision:

**Wrong labelling**. Take English as sample, there may be several sentences $s_{k,l}^{en}|_{l=1}^{L}$ containing the same entity $e_j^{en}$ in the article of $e_i^{en}$. A straightforward solution is to concatenate them into a longer sentence $s_k^{en}$, but this increases the chance to include unrelated sentences.

**Unbalanced information**. Sometimes the pair of sentences convey unbalanced information, e.g., the English sentence in the middle layer (Figure 1) contains *Foust spent 12 seasons in NBA* while the comparable Chinese sentence not.

To address the issues, we propose knowledge attention and cross-lingual attention to filter out unrelated information at sentence level, and at word level respectively.

## 5 Joint Representation Learning

As shown in Figure 2, there are three components in learning cross-lingual word and entity representations, which are trained jointly. In this section, we will describe them in detail.

### 5.1 Mono-lingual Representation Learning

Following Yamada et al. (2016); Cao et al. (2017), we learn mono-lingual word/entity embeddings based on corpus $\mathcal{D}^y$, anchors $\mathcal{A}^y$ and entity network $\mathcal{G}^y$. Capturing the cooccurrence information among words and entities, these embeddings serve as the foundation and will be further extended to bilingual settings using the proposed cross-lingual regularizers, which will be detailed in the next section. Monolingually, we utilize a variant of Skip-gram model (Mikolov et al., 2013c) to predict the contexts given current word/entity:

$$\mathcal{L}_m = \sum_{y \in \{en,zh\}} \sum_{x_i^y \in \{\mathcal{D}^y, \mathcal{A}^y, \mathcal{G}^y\}} \log P(\mathcal{C}(x_i^y)|x_i^y)$$

where $x_i^y$ is either a word or an entity, and $\mathcal{C}(x_i^y)$ denotes: (i) contextual words in a pre-defined window of $x_i^y$ if $x_i^y \in \mathcal{D}^y$, (ii) neighbor entities that linked to $x_i^y$ if $x_i^y \in \mathcal{G}^y$, (iii) contextual words of $w_j^y$ if $x_i^y$ is entity $e_i^y$ in an anchor $\langle w_j^y, e_i^y \rangle \in \mathcal{A}^y$.

### 5.2 Cross-lingual Entity Regularizer

The bilingual EN $\mathcal{G}^{en-zh}$ merges entities in different languages into a unified network, resulting in the possibility of using the same objective as in mono-lingual ENs. Thus, we naturally extend mono-lingual function to cross-lingual settings:

$$\mathcal{L}_e = \sum_{e_i^y \in \{\mathcal{G}^{en-zh}\}} \log P(\mathcal{C}'(e_i^y)|e_i^y)$$

where $\mathcal{C}'(e_i^y)$ denotes cross-lingual contexts—neighbor entities in different languages that linked to $e_i^y$. Thus, by jointly learning mono-lingual representation with cross-lingual entity regularizer, words and entities share more common contexts, and will have similar embeddings. As shown in Figure 1, English entity *NBA* co-occurs with words *basketball* and *player* in texts, so they are embedded closely in the semantic space. Meanwhile, cross-lingual linked entities *NBA* and *NBA (zh)* have similar representations due to the most common neighbor entities, e.g., *Foust*.

### 5.3 Cross-lingual Sentence Regularizer

Comparable sentences provide cross-lingual co-occurrence of words, thus, we can use them to learn similar embeddings for the words that frequently co-occur by minimizing the Euclidean distance as follows:

$$\mathcal{L}_s = \sum_{\langle s_k^{en}, s_{k'}^{zh} \rangle \in \mathcal{S}^{en-zh}} ||\mathbf{s}_k^{en} - \mathbf{s}_{k'}^{zh}||^2$$

where $\mathbf{s}_k^{en}, \mathbf{s}_{k'}^{zh}$ are sentence embeddings. Take English as sample language, we define it as the average sum of word vectors weighted by the combination of two types of attentions:

$$\mathbf{s}_k^{en} = \sum_{l=1}^{L} \psi(e_m^{en}, s_{k,l}^{en}) \sum_{w_i^{en} \in s_{k,l}^{en}} \psi'(w_i^{en}, w_j^{zh}) \mathbf{w}_i^{en}$$

where $s_{k,l}^{en}|_{l=1}^{L}$ are sentences containing the same entity (as mentioned in Section 4.2), and $\psi(e_m^{en}, s_{k,l}^{en})$ is knowledge attention that aims at filtering out wrong labelling sentences, and $\psi'(w_i^{en}, w_j^{zh})$ is cross-lingual attention to deal with the unbalanced information through possible aligned words.

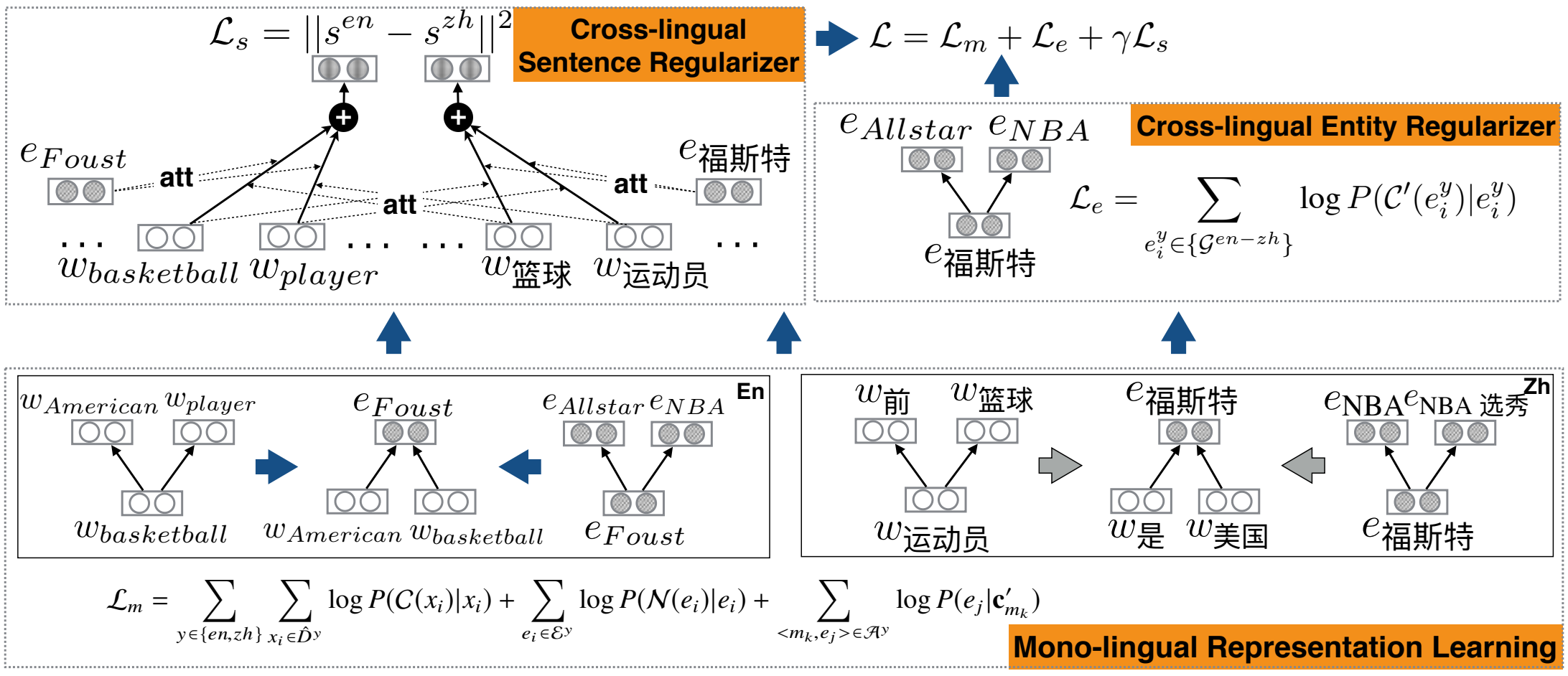

Figure 2: The nerual model for jointly representation learning.

Next, we will introduce the two types of attentions in detail.

### Knowledge Attention

Suppose that sentences $s_{k,l}^{en}|_{l=1}^{L}$ contain the same entities in articles of entity $e_m^{en}$, the wrong labelling errors increase, because some $s_{k,l}^{en}$ is almost irrelevant to $e_m^{en}$. Knowledge attention assigns smaller weights to wrong labelled sentences, and higher weights to related sentences. Thus, we define it proportional to the similarity between $s_{k,l}^{en}$ and $e_m^{en}$:

$$\psi(e_m^{en}, s_{k,l}^{en}) \propto sim(\mathbf{e}_m^{en}, \sum_{w_i^{en}\in s_{k,l}^{en}} \mathbf{w}_i^{en})$$

where $sim$ is similarity measurement. We use cosine similarity in the presented work. Knowledge attention is normalized to satisfy $\sum_l^L \psi(e_m^{en}, s_{k,l}^{en}) = 1$.

### Cross-lingual Attention

Inspired by self-attention mechanism (Lin et al., 2017b), we motivate cross-lingual attention focusing on potential information from comparable sentences themselves. The intuition is to find possible aligned words between languages, and filter out the words without alignments. We define it according to the maximum similarity computed by our cross-lingual word embeddings:

$$\psi'(w_i^{en}, w_j^{zh}) \propto \max_{w_i^{en}\in s_k^{en}, w_j^{zh}\in s_{k'}^{zh}} sim(\mathbf{w}_i^{en}, \mathbf{w}_j^{zh})$$

We set a threshold for discarding non-aligned words if $\psi'(w_i^{en}, w_j^{zh}) < \theta$, and make a normalization for selected words. We set $\theta = 0$ in experiments. Thus, unbalanced information is trimmed to the common meanings between $s_k^{en}$ and $s_{k'}^{zh}$. For example (Figure 1), words *American*, *basketball*, *player* are selected due to their aligned Chinese words 美国, 篮球, 运动员, while *12 seasons* in $s_k^{en}$ or 前 *(former)* in $s_{k'}^{zh}$ are discarded due to low attentions.

The reason of using such regularizer lies in two points: (1) the embeddings of cross-lingual aligned words become closer within the pair of comparable sentences, and meanwhile (2) the distance between their contexts is also minimized, which keeps the same way as used in mono-lingual word embeddings training—the words sharing more contexts have similar embeddings. In this way, our regularizer follows a similar assumption with (Gouws et al., 2015): *The more frequently two words occur in parallel/comparable sentence pairs, the closer their representation will be.*

## 5.4 Training

All above components are jointly trained using the overall objective function as follows:

$$\mathcal{L} = \mathcal{L}_m + \mathcal{L}_e + \gamma\mathcal{L}_s$$

where $\gamma$ is a hyper-parameter to tune the effect of cross-lingual sentence regularizer, and set to 1 in experiments. We use Softmax as probability function, and negative sampling and SGD for efficient optimization (Mikolov et al., 2013a).

## 6 Experiments

In this section, we describe some qualitative analysis with nearest neighbors and quantitative experiments with the tasks of word translation, entity relatedness and cross-lingual entity linking to verify the quality of cross-lingual word embeddings, entity embeddings and the joint inference among them, respectively. The codes of our proposed model can be found in `https://github.com/TaoMiner/MultiLingualEmbedding`.

### 6.1 Experiment Settings

| | Word | | Entity | |
|---|---|---|---|---|
| | vocab (m) | token (b) | vocab (m) | token (b) |
| En | 1.99 | 1.90 | 3.94 | 0.41 |
| Zh | 0.55 | 0.17 | 0.58 | 0.06 |
| Es | 0.70 | 0.48 | 0.70 | 0.04 |
| Ja | 0.46 | 0.45 | 0.88 | 0.08 |
| It | 0.67 | 0.40 | 1.09 | 0.12 |
| Tr | 0.33 | 0.05 | 0.22 | 0.01 |

Table 1: Multi-lingual KB Statistics.

We choose Wikipedia, the April 2017 dump, as multi-lingual KB and six popular languages for evaluation. The preprocessing consists of following steps: converting texts into lower cases, filtering out symbols and low frequency words and entities (less than 5), and tokenizing Chinese corpus using Jieba[4] and Japanese corpus using mecab[5]. The statistics is listed in Table 1. For brevity, we adopt two-letter abbreviations: ‘En’, ‘Zh’, ‘Es’, ‘Ja’, ‘It’ and ‘Tr’ for English, Chinese, Spanish, Japanese, Italian and Turkish, respectively. The token sub-column denotes the total number of word/entity in the entire training corpus, and we use ‘m’ to denote million and ‘b’ for billion.

For cross-lingual settings, we choose five language pairs to compare with state-of-the-art methods, whose statistics is listed in Table 2.

We trained our method using the suggested parameters in Skip-gram model (Mikolov et al., 2013c) and evaluate the embeddings shared by all tasks for fairly comparison. We set training epoch as 2 to ensure convergence, which costs nearly 20 hours on the server with 64 core CPU and 188GB memory. The embedding dimension is set to 200 and context window size is 5. For each positive example, we sample 5 negative examples.

[4]`https://github.com/fxsjy/jieba`
[5]`http://taku910.github.io/mecab/`

| | Cross-lingual Links (m) | Comparable Sentences(m) | Bilingual EN | |
|---|---|---|---|---|
| | | | $\mathcal{E}$(m) | $\mathcal{R}$(b) |
| Es-En | 0.82 | 4.66 | 4.64 | 0.58 |
| Zh-En | 0.51 | 2.02 | 4.52 | 0.57 |
| Ja-Zh | 0.26 | 1.04 | 1.46 | 0.19 |
| It-En | 0.74 | 3.83 | 5.03 | 0.68 |
| Tr-En | 0.15 | 0.75 | 4.16 | 0.44 |

Table 2: Cross-lingual Data Statistics.

### 6.2 Qualitative Analysis

| **Translation words (Chinese)** |
|---|
| 篮球 (+), 篮球队 (basketball team), 湖人 (lakers), 男子篮球 (men’s basketball), 湖人队 (the lakers), 国王队 (the Kings), 美式足球 (American football), 中锋 (center) |
| **Nearest entities (Chinese)** |
| NBA, 篮球 (Basketball) , 控球后卫 (Point guard), NBA 选秀 (draft), 香港男子甲一组男子篮球联赛 (Hong Kong men’s top basketball league), 橄榄球 (American football), 东方篮球队 (Eastern basketball team) |
| **Nearest words** |
| nba, wnba, player, twyman, professional, pick, 76ers |
| **Nearest entities** |
| Professional sports, Varsity letter, Sports agent, All-America, Final four, All-star, College basketball |

Table 3: Cross-lingual nearest words and entities of English word *basketball*.

We manually checked nearest neighbors to have a straightforward impression of the quality of our embeddings. The nearest neighbors of English word *basketball* is listed in Table 3.

As Table 3 shows, we find the correct translation ranked at top 1 (marked by +), and the listed words as well as English nearest words are all basketball related, indicating a higher quality of our cross-lingual word embeddings. Interestingly, we found that although all nearest entities are sports related, e.g., *NBA* or *Professional sports*, there is an obvious culture divergence between Chinese entities and English entities, such as *Hong Kong basketball league* v.s. *All-America*.

### 6.3 Word Translation

Following (Zhang et al., 2017b), we test our cross-lingual word embeddings on benchmark dataset including over 2,000 bilingual word pairs on average. The ground truth is obtained from Open

| | Es-En | | It-En | | Ja-Zh | | Tr-En | | Zh-En | |
|---|---|---|---|---|---|---|---|---|---|---|
| | large | small | large | small | large | small | large | small | large | small |
| TM | - | 48.61 | - | 37.95 | - | 26.67 | - | 11.15 | 4.79 | 21.79 |
| IA | - | 60.41 | - | 46.52 | - | 36.35 | - | 17.11 | 7.08 | 32.29 |
| Bilbowa | 53 | 65.96 | - | - | - | - | - | - | - | - |
| BWESG | 48.88 | 66.38 | 36.84 | 51.29 | 30.93 | 37.80 | 21.36 | 35.59 | 20.57 | 29.17 |
| Adversarial | - | 71.97 | - | 58.60 | - | 43.02 | - | 17.18 | 7.92 | 43.31 |
| Ours-noatt | 68.34 | 77.1 | 62.22 | 65.90 | 37.00 | 42.30 | **57.47** | **60.51** | **35.90** | 42.80 |
| Ours | **70.41** | **78.50** | **63.07** | **67.85** | **41.30** | **46.70** | 54.40 | 59.31 | 35.66 | **44.67** |

Table 4: Word Translation.

Multilingual WordNet[6] or Google translation. We compare all methods using the same vocabulary, and analyze the vocabulary size's impact by setting a nearly 5k small scale and 50k large scale.

We choose several state-of-the-art methods as baseline, using different level of parallel data: (1) TM (Mikolov et al., 2013b), IA (Zhang et al., 2016) are pioneers and popular transformation based methods using **bilingual lexicon**. (2) Bilbowa (Gouws et al., 2015) is typical work using **parallel sentences** and performs quite well. (3) BWESG (Vulic and Moens, 2016) is similar to our method and achieves best performance in the literature of using **comparable data**. (4) Adversarial model (Zhang et al., 2017b) is the state-of-the-arts **without parallel data**. Besides, we remove attention from our method to investigate the impacts from attention mechanisms, marked with *Ours-noatt*.

For fair comparison, we report the results in original paper (Zhang et al., 2017b) except Bilbowa and BWESG, which didn't report their results on the same benchmark datasets. So, we carefully implement them using released codes on the same training corpus as ours with suggested parameters. Nevertheless, we do not have performance reports of Zh-En, It-En, Tr-En and Ja-Zh with Bilbowa due to the lack of parallel data used in the original paper. As shown in Table 4, we can see:

- Our proposed method significantly outperforms all the baseline methods with average gains of 21% and 9.1% on large and small vocabulary. This proves the high quality of our generated cross-lingual data and the effectiveness of our joint framework.
- The pair of languages have similar culture achieves better performance (Es-En, It-En, Tr-En, Ja-Zh) than that have different cultural origins, e.g., Zh-En.
- Languages with richer corpus have better translations because adequate training data helps to capture more accurate cross-lingual semantics (Es-En, It-En, Tr-En v.s. Ja-Zh).
- Our method has less performance reduction between small and large vocabulary than methods based on parallel word pairs, because we adopt a consistent objective function which aligns cross-lingual semantics, and simultaneously keeps their own monolingual semantics.
- Attention mechanisms further improve the performance, mainly because they help to select the most informative words and sentences, filtering out unrelated data.

[6] http://compling.hss.ntu.edu.sg/omw

### 6.4 Entity Relatedness

With respect to our entity embeddings, we have conducted experiments to evaluate English entity relatedness following (Ganea and Hofmann, 2017; Hoffart et al., 2011), in which the dataset contains 3,314 entities, and each entity has 91 candidate entities labeled with 1 or 0, indicating whether they are semantically related. Given an entity, we rank candidate entities according to their similarity based on our embeddings, and evaluate the ranking quality through two standard metrics: normalized discounted cumulative gain (NDCG) (Järvelin and Kekäläinen, 2002) and mean average precision (MAP) (Manning et al., 2008).

To give a comprehensive fair comparison, we choose several widely used and state-of-the-art methods as our baselines, and compare with the results in the original papers: (1) WLM (Milne and Witten, 2008), the popular semantic similarity measurement based on Wikipedia anchor links. (2) ALIGN (Yamada et al., 2016) and MPME (Cao et al., 2017), state-of-the-arts that jointly learn word and entity embeddings using mono-lingual EN. (3) Deep Joint (DJ) model (Ganea and Hofmann, 2017), deep neural model that achieves the best performance of entity relatedness.

| | NDCG | | | MAP |
|---|---|---|---|---|
| | @1 | @5 | @10 | |
| WLM | .54 | .52 | .55 | .48 |
| ALIGN (d=500) | .59 | .56 | .59 | .52 |
| MPME | .61 | .61 | .65 | .58 |
| DJ (d=300) | .63 | .61 | .64 | .58 |
| Ours (Zh-En) | .62 | .62 | .66 | .59 |
| Ours (Es-En) | .61 | .61 | .65 | .59 |
| Ours (Tr-En) | .62 | .62 | .65 | .59 |
| Ours (It-En) | .61 | .61 | .65 | .58 |
| Ours-e (Es-En) | .62 | .62 | .67 | .61 |
| Ours-e (Es-En,epoch=5) | **.64** | **.64** | **.68** | **.62** |

Table 5: Entity Relatedness.

Table 5 shows the results of baseline methods as well as our methods based on different languages. We also test the cases of our method without training cross-lingual words, marked as Ours-e. We can see our method outperforms all baseline methods by introducing cross-lingual information, and all bilingual ENs lead to similar results. Strangely, ALIGN and DJ with more embedding dimensions seemly fails to capture overall relatedness (performance reduction from top@1 to top@5). The best performance of Ours-e implies that training cross-lingual word slightly harms the performance of entity embeddings. We can introduce additional sense embeddings in future (Cao et al., 2017).

Although favorable improvements has been achieved by using our English entity embeddings, it shall be fewer than that of other languages, because resources of English are already quite rich, and even richer than many other languages, thus contributions from other languages will be less significant than vice versa. Due to the limitation of the publication, we neglect to report experiment results on the vice versa direction.

### 6.5 Cross-lingual Entity Linking

Entity linking, the task of identifing the language-specific reference entity for mentions in texts, raises the key challenges of comparing the relevance between entities and contextual words around the mentions (Cao et al., 2015; Nguyen et al., 2016). Recently, the surge of cross-lingual analysis pushes the entity linking task on cross-lingual settings (Ji et al., 2015). Therefore, we comprehensively measure our joint inference ability among words and entities using the tri-lingual EL benchmark dataset KBP2015, which consists of 944 documents and 38,831 mentions, and divides them into 444 and 500 documents for training and evaluation. Note that the main purpose of it is not to beat other EL models but to evaluate the quality of our embeddings, so we adopt a simple classifier GBRT (Gradient Boost Regression Tree) based method as in (Cao et al., 2017; Yamada et al., 2016), replace with our cross-lingual embeddings, and filter out mentions that are out of our vocabulary.

| | English | Spanish | Chinese |
|---|---|---|---|
| Top system | 73.7 | 80.4 | 83.1 |
| Second system | 66.2 | 71.5 | 78.1 |
| Ours | 73.9 | 79.1 | 81.3 |

Table 6: Tri-lingual Entity Linking.

Table 6 shows the top 1 linking accuracy (%). We can see our method performs much better than the second ranked system, and is competitive with the top ranked system. Considering that the systems utilize additional translation tools (Ji et al., 2015), we conclude that our embeddings are high qualified for joint inference among entities and words in different languages.

## 7 Conclusions

In this paper, we propose a novel method to jointly learn cross-lingual word and entity representations that enables effective inference among cross-lingual knowledge bases and texts. Instead of parallel data, we use distant supervision over multi-lingual KB to generate high quality comparable data as cross-lingual supervision signals for two types of regularizer. We introduce attention mechanism to further improve the training quality. A series of experiments on several tasks verify the effectiveness of our methods as well as the quality of cross-lingual word and entity embeddings.

In the future, we will enrich semantics of low-resourced languages by cross-lingual linking to rich-resourced languages, and extend more cross-lingual words and entities to multi-lingual settings.

## 8 Acknowledgments

The work is supported by NSFC key project (No. 61533018, U1736204, 61661146007), Ministry of Education and China Mobile Research Fund (No. 20181770250), and THUNUS NExT++ Co-Lab. Partial financial support from P3ML project funded by BMBF of Germany under grant number 01/S17064 is greatly acknowledged.

## References

Waleed Ammar, George Mulcaire, Yulia Tsvetkov, Guillaume Lample, Chris Dyer, and Noah A. Smith. 2016. Massively multilingual word embeddings. *CoRR*.

Mikel Artetxe, Gorka Labaka, and Eneko Agirre. 2017. Learning bilingual word embeddings with (almost) no bilingual data. In *ACL*.

Antonio Valerio Miceli Barone. 2016. Towards cross-lingual distributed representations without parallel text trained with adversarial autoencoders. In *Rep4NLP@ACL*.

José Camacho-Collados, Mohammad Taher Pilehvar, and Roberto Navigli. 2015. Nasari: a novel approach to a semantically-aware representation of items. In *HLT-NAACL*.

Hailong Cao, Tiejun Zhao, Shu Zhang, and Yao Meng. 2016. A distribution-based model to learn bilingual word embeddings. In *COLING*.

Yixin Cao, Lei Hou, Juanzi Li, and Zhiyuan Liu. 2018. Neural collective entity linking. In *COLING*.

Yixin Cao, Lifu Huang, Heng Ji, Xu Chen, and Juan-Zi Li. 2017. Bridge text and knowledge by learning multi-prototype entity mention embedding. In *ACL*.

Yixin Cao, Juanzi Li, Xiaofei Guo, Shuanhu Bai, Heng Ji, and Jie Tang. 2015. Name list only? target entity disambiguation in short texts. In *EMNLP*.

A. P. Sarath Chandar, Stanislas Lauly, Hugo Larochelle, Mitesh M. Khapra, Balaraman Ravindran, Vikas C. Raykar, and Amrita Saha. 2014. An autoencoder approach to learning bilingual word representations. In *NIPS*.

Meiping Dong, Yong Cheng, Yang Liu, Jia Xu, Maosong Sun, Tatsuya Izuha, and Jie Hao. 2014. Query lattice for translation retrieval. In *COLING*.

Manaal Faruqui and Chris Dyer. 2014. Improving vector space word representations using multilingual correlation. In *EACL*.

Octavian-Eugen Ganea and Thomas Hofmann. 2017. Deep joint entity disambiguation with local neural attention. In *EMNLP*.

Stephan Gouws, Yoshua Bengio, and Gregory S. Corrado. 2015. Bilbowa: Fast bilingual distributed representations without word alignments. In *ICML*.

Xu Han, Zhiyuan Liu, and Maosong Sun. 2016. Joint representation learning of text and knowledge for knowledge graph completion. *CoRR*.

Yanchao Hao, Yuanzhe Zhang, Kang Liu, Shizhu He, Zhanyi Liu, Hua Wu, and Jun Zhao. 2017. An end-to-end model for question answering over knowledge base with cross-attention combining global knowledge. In *ACL*.

Shizhu He, Cao Liu, Kang Liu, and Jun Zhao. 2017. Generating natural answers by incorporating copying and retrieving mechanisms in sequence-to-sequence learning. In *ACL*.

Karl Moritz Hermann and Phil Blunsom. 2014. Multilingual models for compositional distributed semantics. In *ACL*.

Johannes Hoffart, Mohamed Amir Yosef, Ilaria Bordino, Hagen Fürstenau, Manfred Pinkal, Marc Spaniol, Bilyana Taneva, Stefan Thater, and Gerhard Weikum. 2011. Robust disambiguation of named entities in text. In *EMNLP*.

Raphael Hoffmann, Congle Zhang, Xiao Ling, Luke S. Zettlemoyer, and Daniel S. Weld. 2011. Knowledge-based weak supervision for information extraction of overlapping relations. In *ACL*.

Kalervo Järvelin and Jaana Kekäläinen. 2002. Cumulated gain-based evaluation of ir techniques. *TOIS*.

Heng Ji, Joel Nothman, Hoa Trang Dang, and Sydney Informatics Hub. 2016. Overview of tac-kbp2016 tri-lingual edl and its impact on end-to-end cold-start kbp. In *TAC*.

Heng Ji, Joel Nothman, Ben Hachey, and Radu Florian. 2015. Overview of tac-kbp2015 tri-lingual entity discovery and linking. In *TAC*.

Alexandre Klementiev, Ivan Titov, and Binod Bhattarai. 2012. Inducing crosslingual distributed representations of words. In *COLING*.

Tomás Kociský, Karl Moritz Hermann, and Phil Blunsom. 2014. Learning bilingual word representations by marginalizing alignments. In *ACL*.

Guillaume Lample, Alexis Conneau, Ludovic Denoyer, Hervé Jégou, et al. 2018. Word translation without parallel data. *ICLR*.

Yankai Lin, Zhiyuan Liu, and Maosong Sun. 2017a. Neural relation extraction with multi-lingual attention. In *ACL*.

Zhouhan Lin, Minwei Feng, Cícero Nogueira dos Santos, Mo Yu, Bing Xiang, Bowen Zhou, and Yoshua Bengio. 2017b. A structured self-attentive sentence embedding. *CoRR*.

Thang Luong, Hieu Pham, and Christopher D. Manning. 2015. Bilingual word representations with monolingual quality in mind. In *VS@HLT-NAACL*.

Christopher D. Manning, Prabhakar Raghavan, and Hinrich Schütze. 2008. Introduction to information retrieval.

Tomas Mikolov, Kai Chen, Gregory S. Corrado, and Jeffrey Dean. 2013a. Efficient estimation of word representations in vector space. *CoRR*.

Tomas Mikolov, Quoc V Le, and Ilya Sutskever. 2013b. Exploiting similarities among languages for machine translation. *CoRR*.

Tomas Mikolov, Ilya Sutskever, Kai Chen, Gregory S. Corrado, and Jeffrey Dean. 2013c. Distributed representations of words and phrases and their compositionality. In *NIPS*.

David Milne and Ian H. Witten. 2008. An effective, low-cost measure of semantic relatedness obtained from wikipedia links. In *AAAI*.

Mike Mintz, Steven Bills, Rion Snow, and Daniel Jurafsky. 2009. Distant supervision for relation extraction without labeled data. In *ACL/IJCNLP*.

Aditya Mogadala and Achim Rettinger. 2016. Bilingual word embeddings from parallel and non-parallel corpora for cross-language text classification. In *HLT-NAACL*.

Thien Huu Nguyen, Nicolas Fauceglia, Mariano Rodriguez Muro, Oktie Hassanzadeh, Alfio Massimiliano Gliozzo, and Mohammad Sadoghi. 2016. Joint learning of local and global features for entity linking via neural networks. In *COLING*.

Sebastian Ruder, Ivan Vulić, and Anders Søgaard. 2017. A survey of cross-lingual word embedding models. *arXiv preprint arXiv:1706.04902*.

Tianze Shi, Zhiyuan Liu, Yang Liu, and Maosong Sun. 2015. Learning cross-lingual word embeddings via matrix co-factorization. In *ACL*.

Radu Soricut and Nan Ding. 2016. Multilingual word embeddings using multigraphs. *CoRR*.

Mihai Surdeanu, Julie Tibshirani, Ramesh Nallapati, and Christopher D. Manning. 2012. Multi-instance multi-label learning for relation extraction. In *EMNLP-CoNLL*.

Kristina Toutanova, Danqi Chen, Patrick Pantel, Hoifung Poon, Pallavi Choudhury, and Michael Gamon. 2015. Representing text for joint embedding of text and knowledge bases. In *EMNLP*.

Ivan Vulic and Marie-Francine Moens. 2015. Bilingual word embeddings from non-parallel document-aligned data applied to bilingual lexicon induction. In *ACL*.

Ivan Vulic and Marie-Francine Moens. 2016. Bilingual distributed word representations from document-aligned comparable data. *JAIR*.

Zhen Wang, Jianwen Zhang, Jianlin Feng, and Zheng Chen. 2014. Knowledge graph and text jointly embedding. In *EMNLP*.

Zhigang Wang and Juan-Zi Li. 2016. Text-enhanced representation learning for knowledge graph. In *IJCAI*.

Jason Weston, Antoine Bordes, Oksana Yakhnenko, and Nicolas Usunier. 2013a. Connecting language and knowledge bases with embedding models for relation extraction. In *ACL*.

Jason Weston, Antoine Bordes, Oksana Yakhnenko, and Nicolas Usunier. 2013b. Connecting language and knowledge bases with embedding models for relation extraction. In *EMNLP*.

Haiyang Wu, Daxiang Dong, Xiaoguang Hu, Dianhai Yu, Wei He, Hua Wu, Haifeng Wang, and Ting Liu. 2014. Improve statistical machine translation with context-sensitive bilingual semantic embedding model. In *EMNLP*.

Jiawei Wu, Ruobing Xie, Zhiyuan Liu, and Maosong Sun. 2016. Knowledge representation via joint learning of sequential text and knowledge graphs. *CoRR*.

Min Xiao and Yuhong Guo. 2014. Distributed word representation learning for cross-lingual dependency parsing. In *CoNLL*.

Ikuya Yamada, Hiroyuki Shindo, Hideaki Takeda, and Yoshiyasu Takefuji. 2016. Joint learning of the embedding of words and entities for named entity disambiguation. In *CoNLL*.

Ikuya Yamada, Hiroyuki Shindo, Hideaki Takeda, and Yoshiyasu Takefuji. 2017. Learning distributed representations of texts and entities from knowledge base. *TACL*.

Bishan Yang and Tom M. Mitchell. 2017. Leveraging knowledge bases in lstms for improving machine reading. In *ACL*.

Daojian Zeng, Kang Liu, Yubo Chen, and Jun Zhao. 2015. Distant supervision for relation extraction via piecewise convolutional neural networks. In *EMNLP*.

Jing Zhang, Yixin Cao, Lei Hou, Juanzi Li, and Hai-Tao Zheng. 2017a. Xlink: an unsupervised bilingual entity linking system. In *Chinese Computational Linguistics and Natural Language Processing Based on Naturally Annotated Big Data*.

Meng Zhang, Yang Liu, Huanbo Luan, and Maosong Sun. 2017b. Adversarial training for unsupervised bilingual lexicon induction. In *ACL*.

Yuan Zhang, David Gaddy, Regina Barzilay, and Tommi S. Jaakkola. 2016. Ten pairs to tag - multilingual pos tagging via coarse mapping between embeddings. In *HLT-NAACL*.

Hao Zhu, Ruobing Xie, Zhiyuan Liu, and Maosong Sun. 2017. Iterative entity alignment via joint knowledge embeddings. In *IJCAI*.

Will Y. Zou, Richard Socher, Daniel M. Cer, and Christopher D. Manning. 2013. Bilingual word embeddings for phrase-based machine translation. In *EMNLP*.